\documentclass{article}
\usepackage{spconf,amsmath,graphicx,hyperref}
\usepackage{multirow}
\usepackage{graphicx}   % 插图宏包
\usepackage{subcaption} % 子图宏包
\usepackage{amssymb}  % for \checkmark
\usepackage{graphicx} % for table width
\usepackage{booktabs} % for better lines
\usepackage{balance}
\usepackage{flushend}

\usepackage{microtype}
\usepackage{multirow}
\usepackage{booktabs}
\usepackage{subcaption}
\usepackage{verbatimbox}
\usepackage[table]{xcolor}
\usepackage{amssymb}% 
\usepackage{dsfont}
\usepackage{pifont}       % \ding{xx}
\usepackage{bbding}       % 
\usepackage{fontawesome}

\definecolor{c1}{HTML}{95bddc}
\definecolor{c2}{HTML}{c2d1e5}
\definecolor{c3}{HTML}{fe793d}
\definecolor{c4}{HTML}{fb4c1f}
\definecolor{c5}{HTML}{b71a3b}
\definecolor{c6}{HTML}{7e0f12}

\title{CG-DMER: Hybrid Contrastive-Generative Framework for Disentangled Multimodal ECG Representation Learning}

\name{Ziwei Niu\textsuperscript{1,3}  \quad Hao Sun\textsuperscript{4}  \quad Shujun Bian\textsuperscript{1} \quad Xihong Yang\textsuperscript{2} \quad Lanfen Lin\textsuperscript{3} \quad Yuxin Liu\textsuperscript{1} \quad Yueming Jin\textsuperscript{1,2 \faEnvelopeO}}

\address{
\textsuperscript{1} Department of Biomedical Engineering, National University of Singapore, Singapore, Singapore\\
\textsuperscript{2} Department of Electrical and Computer Engineering, National University of Singapore, Singapore, Singapore\\
\textsuperscript{3} College of Computer Science and Technology, Zhejiang University, Hangzhou, China\\
\textsuperscript{4} College of information science and engineering, Ritsumeikan university, ibaraki Osaka, Japan}
\begin{document}
\ninept
\maketitle
\begin{abstract}
Accurate interpretation of electrocardiogram (ECG) signals is crucial for diagnosing cardiovascular diseases. Recent multimodal approaches that integrate ECGs with accompanying clinical reports show strong potential, but they still face two main concerns from a modality perspective: (1) intra-modality: existing models process ECGs in a lead-agnostic manner, overlooking spatial–temporal dependencies across leads, which restricts their effectiveness in modeling fine-grained diagnostic patterns; (2) inter-modality: the existing methods directly align ECG signals with clinical reports, introducing modality-specific biases due to the free-text nature of the reports. In light of these two issues, we propose CG-DMER, a contrastive-generative framework for disentangled multimodal ECG representation learning, powered by two appealing designs: (1) Spatial-temporal masked modeling is designed to better capture fine-grained temporal dynamics and inter-lead spatial dependencies by applying masking across both spatial and temporal dimensions and reconstructing the missing information. (2) Representation disentanglement and alignment strategy is designed to further mitigate unnecessary noise and modality-specific biases by designing modality-specific and modality-shared encoders, which ensures a clearer separation between modality-invariant and modality-specific representations. Experiments on three public datasets demonstrate that CG-DMER achieves state-of-the-art performance across diverse downstream tasks. 
\end{abstract}
\vspace{-10pt}
\section{Introduction}
\vspace{-5pt}
\label{sec:intro}
Recent advances in deep learning have achieved remarkable success in leveraging electrocardiograms (ECGs) for the automated classification of cardiovascular diseases (CVD). %However, most methods are supervised and require large amounts of annotated data, which is costly and requires extensive expert effort in annotation.
However, most methods are supervised and depend on large, expertly annotated datasets, which are costly to obtain.
To mitigate these limitations, self-supervised learning (SSL) has emerged as a compelling alternative, enabling models to learn generalizable representations directly from unlabeled data. 
These representations can then be fine-tuned for specific downstream tasks, reducing dependence on large labeled datasets. Building on these advancements, ECG self-supervised learning (eSSL) has become a promising paradigm for extracting robust features from large-scale unlabeled ECGs. Contrastive eSSL (C-eSSL)~\cite{chen2020simple,mocov3,astcl,kiyasseh2021clocs,mckeen2024ecg} leverages contrastive learning techniques to capture clinically relevant patterns without requiring labeled data, thereby facilitating the differentiation of various physiological states. Generative eSSL (G-eSSL)~\cite{hu2023spatiotemporal,zhang2022maefe,crt,na2024guiding} focuses on modeling the underlying distribution of ECG signals by learning to reconstruct, predict, or generate physiologically realistic waveforms from unlabeled data. By leveraging these techniques, eSSL enhances model generalization and improves performance across various downstream clinical applications.

Despite these advancements, existing eSSL approaches have predominantly focused on learning representations from raw ECG waveforms, often overlooking the rich diagnostic and contextual information embedded within clinical text reports. %These reports, which contain expert interpretations, annotations, and detailed descriptions of cardiac conditions, offer complementary insights that can enhance the understanding of ECG signals. 
As a result, this oversight along with the requirement for annotated samples in downstream tasks limits the versatility of eSSL. Until recently, multimodal approaches, including ETP~\cite{liu2024etp}, MERL~\cite{liu2024zero} and C-MELT~\cite{c-melt} have attempted to address these limitations. Despite their great success, they still face two main challenges from a modality-wise perspective: \textbf{\textit{{i}) Intra-modality issue}}: Existing methods encode ECG signals in a lead-agnostic manner, overlooking the unique spatial and temporal characteristics of individual ECG leads, which limits their effectiveness in capturing fine-grained diagnostic information. \textbf{\textit{{ii})  Inter-modality issue}}: Recent methods directly align ECG signals with clinical reports, introducing unnecessary noise and modality-specific biases due to the free-text nature of the reports.

Driven by the above analysis, we propose CG-DMER, a simple yet effective 
\underline{C}ontrastive-\underline{G}enerative framework for \underline{D}isentangled \underline{M}ultimodal \underline{E}CG \underline{R}epresentation Learning. The core of CG-DMER lies in capturing fine-grained input details and enhancing ECG-text feature discrimination. Specifically, we propose a spatial-temporal masked modeling scheme for ECG signals, which applies masking across both lead-wise and temporal dimensions to encourage the model to learn fine-grained temporal dynamics and inter-lead spatial dependencies. In parallel, we apply masked reconstruction to clinical text reports, facilitating the learning of rich semantic representations. To further mitigate unnecessary noise and modality-specific biases, we design modality-specific and modality-shared encoders to disentangle features, promoting a clearer separation between modality-invariant and modality-specific representations. Cross-modal contrastive learning is then applied to the shared ECG-text representations, enhancing feature discrimination and strengthening cross-modal alignment. Through this unified framework, our method facilitates the learning of generalizable and semantically rich multimodal representations.
%, leading to improved performance across various downstream clinical tasks.

\begin{figure*}[t]
\centering
    \includegraphics[width=0.97\linewidth]{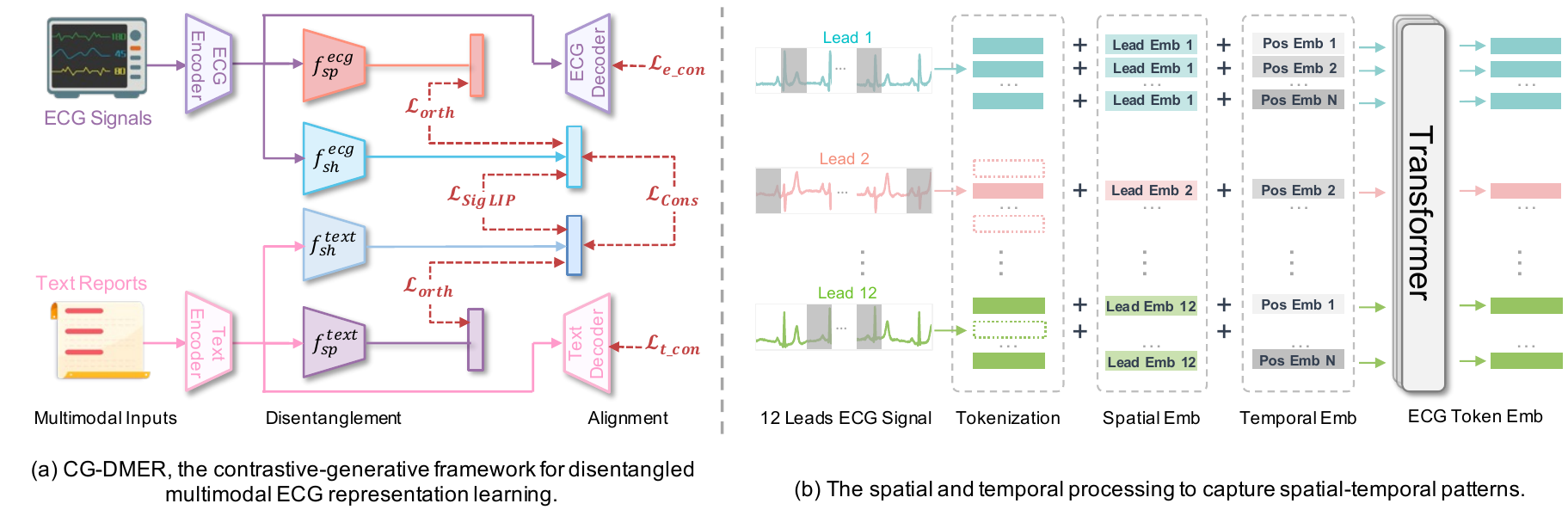} 
    \caption{A schematic view of the proposed method.}
    \vspace{-5pt}
    \label{fig2} 
\end{figure*}

\vspace{-10pt}
\section{Method}
\vspace{-5pt}
% \subsection{Overview}
In multimodal ECG representation learning, given a training data set $\mathcal{X}$ consisting of pairs of $\textit{N}$ ECG and text reports, we represent each pair as (${E}_i \in \mathbb{R}^{L \times T}$, ${R}_i \in \mathcal{V^*}$), where $L$ and $T$ denote the lead and length of the raw ECG records ${E}_i$, and ${R}_i$ denotes the associated text report, respectively, with $i = 1, 2, 3, ..., N$. Our CG-DMER framework learns general representations directly from ECG signals and the corresponding text reports as Figure~\ref{fig2} describes. 
\vspace{-5pt}
\subsection{Multimodal Masked Modeling}\label{sec3.2}
Masked modeling is a representation learning technique that reconstructs all or part of the input data, as demonstrated in frameworks such as masked image modeling and Masked Autoencoder (MAE)~\cite{MAE}. This approach offers a powerful means of jointly learning robust ECG-text representations by leveraging the underlying cross-modal data distribution and reconstructing missing segments to capture fine-grained dependencies. Accordingly, we consider a pretext task that involves reconstructing randomly masked portions of both ECG signals and associated clinical text reports.

\noindent \textbf{Spatial–Temporal Masked ECG Modeling.}  
An $L$-lead ECG records cardiac activity from multiple spatial viewpoints, making both spatial and temporal information critical. We propose a spatial–temporal masked ECG modeling approach that applies masking across temporal and inter-lead dimensions, enabling the model to capture correlations and fine-grained patterns.  

Specifically, ECG signal ${E}_i \in \mathbb{R}^{L \times T}$ is divided into $N$ temporal patches along the temporal axis, yielding $L \times N$ patches $Patch^l =\left\{\right.$Patch $_1^l, \ldots$, Patch $\left._N^l\right\} \in \mathbb{R}^{T}$. Each patch is passed through a series of convolutional layers, followed by GELU~\cite{hendrycks2016gaussian} activation and group normalization, to perform tokenization. To capture spatial context, we introduce lead embeddings $[spa_1,\ldots,spa_{L}]$, with $spa_l \in \mathbb{R}^d$. 
% The final input is the sum of token and lead (spatial) embeddings:
% \begin{equation}
% \begin{aligned}
% & spa_1+\mathbf{W} E_i^l\left[p_1\right], \ldots, spa_1+\mathbf{W} E_i^l\left[p_N\right], \ldots, \\
% & spa_{L}+\mathbf{W} E_i^l\left[p_1\right], \ldots, spa_{L}+\mathbf{W} E_i^l\left[p_N\right].
% \end{aligned}
% \end{equation}
To preserve lead-specific characteristics, we tokenize each lead independently. Let $E_i^l[p_n]$ denote the token corresponding to the $n$-th temporal segment of the $l$-th lead. Unlike MERL~\cite{liu2024zero} and C-MERL~\cite{c-melt}, which
generate a single token for each temporal segment of a $L$-lead ECG, our method produces separate tokens per lead. To retain temporal ordering, we add shared temporal (positional) embeddings $[\text{temp}_1,\ldots,\text{temp}_N]$, allowing consistency across leads while preserving spatial specificity. The final input embeddings are constructed by summing the token embeddings, the corresponding lead-specific spatial embeddings, and the shared temporal embeddings as follows:
\begin{equation}
\begin{aligned}
& \operatorname{temp}_1+\operatorname{spa}_1+\mathbf{W} E_i^l\left[p_1\right] \quad \ldots, \quad \operatorname{temp}_N+\operatorname{spa}_1+\mathbf{W} E_i^l\left[p_N\right] \ldots, \quad \\
& \operatorname{temp}_1+\operatorname{spa}_{L}+\mathbf{W} E_i^l\left[p_1\right]  \ldots, \quad \operatorname{temp}_N+\operatorname{spa}_{L}+\mathbf{W} E_i^l\left[p_N\right] .
\end{aligned}
\end{equation}

We randomly select a set of indices $\mathcal{M}_j$ to mask for each sample $j$ with ratio $r$. For $N$ patches per lead, the total number of masked patches is $|\mathcal{M}_j| = N \times L \times r$. Masking is applied uniformly across leads. The encoder receives unmasked embeddings $\{Z_i\}_{i \notin \mathcal{M}_j}$, which are processed by eight Transformer layers with multi-head self-attention to capture temporal and inter-lead dependencies.  

During decoding, the full sequence (unmasked embeddings + shared mask embeddings) is fed to reconstruct the masked patches. The reconstruction objective is:
\begin{equation}
\mathcal{L}_{e\_rec} = \frac{1}{B} \sum_{j=1}^{B} \frac{1}{|\mathcal{M}_j|} \sum_{i \in \mathcal{M}_j} \bigl\| \widehat{Patch}_{j,i} - Patch_{j,i} \bigr\|_2^2.
\end{equation}

% \begin{equation}
% \begin{aligned}
% \mathcal{L}_{e\_rec} = \frac{1}{B}\frac{1}{|\mathcal{M}|} \sum_{j=1}^{B}\sum_{i=1}^{\mathcal{M}} \left\| \left\{\widehat{Patch}_i\right\}_{i \in \mathcal{M}_j} - \left\{Patch_i\right\}_{i \in \mathcal{M}_j} \right\|_2^2
% \end{aligned}
% \end{equation}

\noindent \textbf{Masked Text Report Reconstruction.}  
Following MERL~\cite{liu2024zero}, we adopt Med-CPT~\cite{jin2023medcpt} as the text encoder. 
Text reports are tokenized, and 15\% of tokens are randomly masked, and the set of masked indices is denoted as $\mathcal{M}_j$. 
The corrupted sequence $\mathbf{t}_{j \setminus \mathcal{M}_j}$, where masked tokens are replaced with sentinel tokens, is fed into the encoder. 
The decoder autoregressively predicts the masked tokens, and the training objective is the cross-entropy loss:
\begin{equation}
\mathcal{L}_{t\_rec} = -\frac{1}{B} \sum_{j=1}^{B} \frac{1}{|\mathcal{M}_j|} \sum_{m \in \mathcal{M}_j} \log P \bigl(t_{j,m} \mid \mathbf{t}_{j \setminus \mathcal{M}_j} ; \theta \bigr),
\end{equation}
where $t_{j,m}$ is the ground-truth token at position $m$, $\mathbf{t}_{j \setminus \mathcal{M}_j}$ is the masked input sequence, and $\theta$ denotes model parameters.

\vspace{-5pt}
\subsection{Representation Disentanglement and Alignment}\label{sec3.3}
\textbf{Representation Disentanglement:}  
Previous alignment strategies for multimodal ECG pretraining often attempt to directly align representations via contrastive learning. However, this can be suboptimal because of intrinsic differences between modalities, such as data type, training paradigm, and semantic context. Simply forcing modality-specific features into a shared space may create incompatibilities and negatively affect downstream performance. To overcome this, we introduce a disentanglement strategy that separates representations into modality-specific and modality-shared components for both ECG signals and text reports. This design explicitly preserves unique modality information while isolating common semantic features, leading to more stable integration.  

Based on the representations $\mathbf{E}^{ecg}$ and $\mathbf{E}^{text}$ obtained from the ECG and text encoders, respectively, we disentangle them into:
\begin{equation}
\begin{aligned}
& \mathbf{h}_{s p}^{ecg}=f_{s p}^{ecg}\left(\mathbf{E}^{ecg}\right), \mathbf{h}_{s h}^{ecg}=f_{s h}^{ecg}\left(\mathbf{E}^{ecg}\right), \\
& \mathbf{h}_{s p}^{text}=f_{s p}^{text}\left(\mathbf{E}^{text}\right), \mathbf{h}_{s h}^{text}=f_{s h}^{text}\left(\mathbf{E}^{text}\right),
\end{aligned}
\end{equation}
where $f_{sh}(\cdot)$ and $f_{sp}(\cdot)$ denote encoder networks for shared and specific parts, implemented as MLPs. Since modality-shared and modality-specific components are not naturally separated, we apply an orthogonality constraint to reduce redundancy by minimizing cosine similarity between $h_{sh}$ and $h_{sp}$:
\begin{equation}
\mathcal{L}_{orth}=\frac{1}{B} \sum_{i=1}^B\!\left(\mathbf{S}(\mathbf{h}_{s p_i}^{text}, \mathbf{h}_{s h_i}^{text})^2+\mathbf{S}(\mathbf{h}_{s p_i}^{ecg}, \mathbf{h}_{s h_i}^{ecg})^2\right),
\end{equation}
where $\mathbf{S}$ is cosine similarity and $B$ is the batch size.  

\noindent
\textbf{Representation Alignment:}  
Medical reports provide detailed descriptions that correspond to ECG signals, so paired samples should naturally align in the multimodal feature space. Yet in masked autoencoder architectures, generative reconstruction objectives and discriminative contrastive objectives often conflict, which can hinder embedding quality. To alleviate this, following C-MERL~\cite{c-melt}, we adopt the SigLIP loss~\cite{zhai2023sigmoid}, originally proposed for image–text pairs, and extend it to ECG–text alignment:
\begin{equation}
\mathcal{L}_{\text {SigLIP}}=-\frac{1}{B} \sum_{i=1}^{B} \sum_{j=1}^{B} \log \left(\frac{1}{1+e^{-y_{i j} {\mathbf{h}_{{sh}_i}^{ecg}}^{\top}\mathbf{h}_{{s h}_j}^{text}}}\right),
\end{equation}
where $y_{ij}=1$ for positive ECG–text pairs, and $y_{ij}=-1$ otherwise.  

We then follow~\cite{liu2024etp,liu2024zero} and employ additional contrastive loss functions to maximize the posterior probability of the shared ECG representation $\mathbf{h}_{s h}^{ecg}$ given its paired text representation $\mathbf{h}_{s h}^{text}$. Cosine similarities are defined as $s_{i,i}^{e2t}={\mathbf{h}_{s h}^{ecg}}^{\top} \mathbf{h}_{s h}^{text}$ and $s_{i,i}^{t2e}={\mathbf{h}_{s h}^{text}}^{\top}\mathbf{h}_{s h}^{ecg}$. The corresponding losses are:
\begin{equation}
\mathcal{L}_{i, j}^{e 2 t}=-\log \frac{\exp \left(s_{i, j}^{e 2 t} / \tau\right)}{\sum_{k=1}^B \mathds{1}_{[k \neq i]} \exp \left(s_{i, k}^{e 2 t} / \tau\right)},
\end{equation}
\begin{equation}
\mathcal{L}_{i, j}^{t 2 e}=-\log \frac{\exp \left(s_{i, j}^{t 2 e} / \tau\right)}{\sum_{k=1}^B \mathds{1}_{[k \neq i]} \exp \left(s_{i, k}^{t 2 e} / \tau\right)}, 
\end{equation}
which are then combined as:
\begin{equation}
\mathcal{L}_{Cons}=\frac{1}{2B} \sum_{i=1}^B \sum_{j=1}^B\left(\mathcal{L}_{i, j}^{e 2 t}+\mathcal{L}_{i, j}^{t 2 e}\right).
\end{equation}
Here $\tau=0.07$ is the temperature hyper-parameter and $B$ the batch size.  

\noindent \textbf{Overall Objective.}  
Our model learns representative ECG features by jointly minimizing $\mathcal{L}_{Cons}$, $\mathcal{L}_{t\_rec}$, $\mathcal{L}_{orth}$ and $\mathcal{L}_{SigLIP}$. The overall training objective is:
\begin{equation}
\mathcal{L}_{Full}= \mathcal{L}_{Cons} + \lambda_1 \mathcal{L}_{t\_rec} + \lambda_2 \mathcal{L}_{orth} + \lambda_3 \mathcal{L}_{SigLIP},
\end{equation}
where $\lambda_1$, $\lambda_2$, and $\lambda_3$ are trade-off weights.

\section{Experiments}
\label{sec:pagestyle}
\subsection{Pre-training Configurations}
\textbf{Pre-train Dataset:} We utilize the MIMIC-IV-ECG database~\cite{gow2023mimic} for pre-training, which comprises 800,035 paired samples from 161,352 unique subjects. The dataset includes multiple 10-second ECG recordings sampled at 500 Hz, along with corresponding textual reports. To ensure a fair comparison with existing methods, we follow the preprocessing protocol of the MERL~\cite{liu2024zero} framework. After preprocessing, we obtain a total of 779,891 samples for model training. 

{\setlength{\parindent}{0pt}
\textbf{Pre-train Implementation:}  The AdamW optimizer is used with a learning rate of \(2 \times 10^{-4}\) and a weight decay of \(1 \times 10^{-5}\). We pre-train CG-DMER for 50 epochs, applying a cosine annealing scheduler for learning rate adjustments. A batch size of 512 is maintained per GPU, and all experiments are conducted on eight NVIDIA RTX A5000.
}

\begin{table*}[t!]
\scriptsize
\centering
\caption{\small{Linear probing performance comparison between CG-DMER and existing ECG representation learning methods. The best and second-best results are highlighted in \textcolor{red}{red} and \textcolor{blue}{blue}, respectively. $\dagger$ denotes results reproduced by our implementation.}}
\vspace{-8pt}
\renewcommand{\arraystretch}{0.95}
\setlength{\tabcolsep}{1.5mm}{  
\begin{tabular}{l|ccc|ccc|ccc|ccc|ccc|ccc}
    \toprule[1.2pt]
     & \multicolumn{3}{c}{PTBXL-Super} & \multicolumn{3}{c}{PTBXL-Sub} 
     & \multicolumn{3}{c}{PTBXL-Form} & \multicolumn{3}{c}{PTBXL-Rhythm} 
     & \multicolumn{3}{c}{CPSC2018} & \multicolumn{3}{c}{CSN} \\
    Method & 1\% & 10\% & 100\% & 1\% & 10\% & 100\% & 1\% & 10\% & 100\% 
           & 1\% & 10\% & 100\% & 1\% & 10\% & 100\% & 1\% & 10\% & 100\% \\
    \midrule[1.2pt]
    Random Init (CNN) & 70.45 & 77.09 & 81.61 & 55.82 & 67.60 & 77.91 & 55.82 & 62.54 & 73.00 & 46.26 & 62.36 & 79.29 & 54.96 & 71.47 & 78.33 & 47.22 & 63.17 & 73.13 \\
    Random Init (Trans) & 70.31 & 75.27 & 77.54 & 53.36 & 67.56 & 77.43 & 53.47 & 61.84 & 72.08 & 45.36 & 60.33 & 77.26 & 52.93 & 68.00 & 77.44 & 45.55 & 60.23 & 71.37 \\
    \midrule
    SimCLR~\cite{simclr} & 63.41 & 69.77 & 73.53 & 60.84 & 68.27 & 73.39 & 54.98 & 56.97 & 62.52 & 51.41 & 69.44 & 77.73 & 59.78 & 68.52 & 76.54 & 59.02 & 67.26 & 73.20 \\
    BYOL~\cite{byol} & 71.70 & 73.83 & 76.45 & 57.16 & 67.44 & 71.64 & 48.73 & 61.63 & 70.82 & 41.99 & 74.40 & 77.17 & 60.88 & 74.42 & 78.75 & 54.20 & 71.92 & 74.69 \\
    BarlowTwins~\cite{zbontar2021barlow} & 72.87 & 75.96 & 78.41 & 62.57 & 70.84 & 74.34 & 52.12 & 60.39 & 66.14 & 50.12 & 73.54 & 77.62 & 55.12 & 72.75 & 78.39 & 60.72 & 71.64 & 77.43 \\
    MoCo-v3~\cite{mocov3} & 73.19 & 76.65 & 78.26 & 55.88 & 69.21 & 76.69 & 50.32 & 63.71 & 71.31 & 51.38 & 71.66 & 74.33 & 62.13 & 76.74 & 75.29 & 54.61 & 74.26 & 77.68 \\
    SimSiam~\cite{simsiam} & 73.15 & 72.70 & 75.63 & 62.52 & 69.31 & 76.38 & 55.16 & 62.91 & 71.31 & 49.30 & 69.47 & 75.92 & 58.35 & 72.89 & 75.31 & 58.25 & 68.61 & 77.41 \\
    TS-TCC~\cite{tstcc} & 70.73 & 75.88 & 78.91 & 53.54 & 66.98 & 77.87 & 48.04 & 61.79 & 71.18 & 43.34 & 69.48 & 78.23 & 57.07 & 73.62 & 78.72 & 55.26 & 68.48 & 76.79 \\
    CLOCS~\cite{clocs} & 68.94 & 73.36 & 76.31 & 57.94 & 72.55 & 76.24 & 51.97 & 57.96 & 72.65 & 47.19 & 71.88 & 76.31 & 59.59 & 77.78 & 77.49 & 54.38 & 71.93 & 76.13 \\
    ASTCL~\cite{astcl} & 72.51 & 77.31 & 81.02 & 61.86 & 68.77 & 76.51 & 44.14 & 60.93 & 66.99 & 52.38 & 71.98 & 76.05 & 57.90 & 77.01 & 79.51 & 56.40 & 70.87 & 75.79 \\
    CRT~\cite{crt} & 69.68 & 78.24 & 77.24 & 61.98 & 70.82 & 78.67 & 46.41 & 59.49 & 68.73 & 47.44 & 73.52 & 74.41 & 58.01 & 76.43 & 82.03 & 56.21 & 73.70 & 78.80 \\
    ST-MEM~\cite{na2024guiding} & 61.12 & 66.87 & 71.36 & 54.12 & 57.86 & 63.59 & 55.71 & 59.99 & 66.07 & 51.12 & 65.44 & 74.85 & 56.69 & 63.32 & 70.39 & 59.77 & 66.87 & 71.36 \\
    \midrule
    
    $\text{ETP}^ \dagger$~\cite{liu2024etp} & 77.25 & 83.06 & 83.90 & 60.34 & 76.66 & 82.07 & 56.60 & 68.75 & 76.13 & 51.88 & 77.32 & 81.00 & 67.67 & 82.40 & 87.97 & 62.05 & 77.30 & 84.44 \\
    MERL (ResNet)~\cite{liu2024zero} & 82.39 & 86.27 & \textcolor{blue}{88.67} & 64.90 & \textcolor{blue}{80.56} & \textcolor{blue}{84.72} & 58.26 & \textcolor{blue}{72.43} & 79.65 & 53.33 & 82.88 & 88.34 & 70.33 & 85.32 & 90.57 & \textcolor{blue}{66.60} & \textcolor{blue}{82.74} & 87.95 \\
    MERL (ViT)~\cite{liu2024zero} & 78.64 & 83.90 & 85.27 & 61.41 & 77.55 & 82.98 & 56.32 & 69.11 & 77.66 & 52.16 & 78.07 & 81.83 & 69.25 & 82.82 & 89.44 & 63.66 & 78.67 & 84.87 \\
    $\text{C-MELT}^ \dagger$~\cite{c-melt} & \textcolor{blue}{83.54} & \textcolor{blue}{87.69} & 88.22 & \textcolor{blue}{67.60} & 80.37 & 83.96 & \textcolor{blue}{59.55} & 71.86 & \textcolor{blue}{80.54} & \textcolor{blue}{62.88} & \textcolor{blue}{83.00} & \textcolor{blue}{88.76} & \textcolor{blue}{70.50} & \textcolor{blue}{86.00} & \textcolor{blue}{90.72} & 66.01 & 82.11 & \textcolor{blue}{91.94} \\
    \midrule
    \rowcolor{gray!20}
    \textbf{CG-DMER (Ours)} & \textcolor{red}{84.22} & \textcolor{red}{88.18} & \textcolor{red}{90.31} & \textcolor{red}{69.60} & \textcolor{red}{81.58} & \textcolor{red}{87.67} & \textcolor{red}{60.25} & \textcolor{red}{73.95} & \textcolor{red}{81.01} & \textcolor{red}{64.47} & \textcolor{red}{84.69} & \textcolor{red}{90.76} & \textcolor{red}{71.70} & \textcolor{red}{86.43} & \textcolor{red}{91.32} & \textcolor{red}{70.70} & \textcolor{red}{83.21} & \textcolor{red}{93.55} \\
\bottomrule[1.2pt]
\end{tabular}
}
\label{tab1}
% \vspace{-5pt}
\end{table*}

\begin{table*}[t]
    \centering
    \scriptsize
    \renewcommand{\arraystretch}{0.95}
    \setlength{\tabcolsep}{4.8mm}{  
    % \scriptsize
    \caption{Zero-shot performance comparison across multiple datasets. The best and second-best results are highlighted in \textcolor{red}{red} and \textcolor{blue}{blue}, respectively. $\dagger$ denotes results reproduced by our implementation.}
    \vspace{-8pt}
    \label{tab2}
    \begin{tabular}{l|ccccccc}
        \toprule[1.2pt]
        Methods & PTBXL-Super & PTBXL-Sub & PTBXL-Form & PTBXL-Rhythm & CPSC2018 & CSN & Average\\
        \midrule[1.2pt]
        $\text{ETP}^ \dagger$~\cite{liu2024etp} & 49.63 & 62.25 & 57.44 & 66.56 & 67.13 & 60.22 & 60.54 \\
        MERL (ResNet)~\cite{liu2024zero} & 74.20 & \textcolor{red}{75.70} & \textcolor{blue}{65.90} & 78.50 & \textcolor{blue}{82.80} & 74.40 & \textcolor{blue}{75.25} \\
        MERL (Vit)~\cite{liu2024zero} & 73.43 & 72.62 & 62.75 & 75.90 & 79.21 & 72.37 & 72.71 \\
        $\text{C-MELT}$~\cite{c-melt} & \textcolor{blue}{75.12} & 74.38 & 65.55 & \textcolor{blue}{79.87} & 81.61 & \textcolor{blue}{74.43} & 75.16 \\
        \rowcolor{gray!20}
        \textbf{CG-DMER (Ours)} & \textcolor{red}{75.65}  & \textcolor{blue}{74.62} & \textcolor{red}{66.44} & \textcolor{red}{80.77}  & \textcolor{red}{82.93} & \textcolor{red}{75.59} & \textcolor{red}{76.00}  \\
        \bottomrule[1.2pt]
    \end{tabular}}
\end{table*}

\subsection{Downstream Tasks Configurations}
We evaluate our framework on two downstream tasks: linear probing and zero-shot classification, using three widely adopted ECG datasets that together cover more than 100 cardiac conditions. The macro AUC is used as the evaluation metric for all tasks.  

{\setlength{\parindent}{0pt}  
\textbf{(1) PTB-XL}~\cite{wagner2020ptb} contains 21,837 ECG recordings from 18,885 patients. Each 12-lead ECG is sampled at 500 Hz for 10 seconds. Following the official annotation protocol, four subsets are provided for multi-label classification: 5 Superclasses, 23 Subclasses, 19 Form classes, and 12 Rhythm classes.  

\textbf{(2) CPSC2018}~\cite{liu2018open} includes 6,877 12-lead ECGs sampled at 500 Hz, with durations ranging from 6 to 60 seconds. Each record is annotated with one of 9 diagnostic categories.  

\textbf{(3) CSN}~\cite{zheng2020optimal, zheng2022large} consists of 45,152 12-lead ECGs sampled at 500 Hz for 10 seconds. After dropping records with ‘unknown’ annotations, the curated version contains 23,026 samples labeled with 38 categories.  
}
\vspace{-5pt}
\subsection{Evaluation on linear probing}
%要把实验结果落到所提出的两个创新点上
Linear probing has become the standard evaluation protocol for ECG self-supervised learning (eSSL), as it fairly measures the quality of learned ECG representations on uni-modal downstream tasks after pretraining. 
Table~\ref{tab1} reports the linear probing results of our proposed CG-DMER against existing eSSL and multimodal methods. 
Across six datasets and training data ratios ranging from 1\% to 100\%, CG-DMER consistently outperforms eSSL baselines. Remarkably, with only 10\% of labeled data, CG-DMER surpasses all eSSL methods trained on 100\% of the data. 
This demonstrates its strong generalization ability, driven by clinical text supervision that yields more discriminative and semantically enriched ECG representations. 
Beyond eSSL, CG-DMER also achieves superior results over multimodal counterparts across all datasets. 
These gains highlight the effectiveness of our hybrid contrastive–generative paradigm, which combines complementary strengths of both objectives. 
Together with the proposed feature disentanglement and alignment strategy, CG-DMER captures fine-grained physiological patterns and modality-specific cues, enabling robust generalization across diverse clinical tasks.
% Linear probing has become the standardized evaluation protocol for ECG self-supervised learning (eSSL) methods, as it provides a consistent and fair assessment of the quality of learned ECG representations on uni-modal downstream tasks after pretraining. 
% Table~\ref{tab1} presents the linear probing results of our proposed CG-DMER, alongside existing eSSL and multimodal learning methods. Our CG-DMER consistently outperforms eSSL baselines across six datasets and a wide range of training data ratios (from 1\% to 100\%). Notably, CG-DMER trained with only 10\% of the labeled data surpasses all eSSL methods that use 100\% of the data on six datasets. This highlights the strong generalization capability of CG-DMER, which benefits significantly from clinical text supervision to learn more discriminative and semantically enriched ECG representations. In addition to eSSL methods, CG-DMER also achieves superior performance compared to other multimodal approaches across all datasets. This demonstrates the effectiveness of our hybrid contrastive-generative learning paradigm, which integrates complementary strengths of both objectives. Combined with the proposed feature decoupling and alignment strategy, CG-DMER is better equipped to capture fine-grained physiological patterns and modality-specific cues, enabling robust generalization across diverse clinical tasks and data distributions.
\vspace{-5pt}
\subsection{Evaluation on zero-shot learning}
Zero-shot classification with text prompts is widely used to evaluate representation quality, as it tests a model’s ability to generalize to unseen tasks without annotated samples. Conventional zero-shot methods rely solely on category names, which often provide limited context for accurate classification. To address this, we incorporate Large Language Models (LLMs) to generate richer, clinically relevant category descriptions. In particular, we use GPT-4o to customize prompts for each category.  
We compare CG-DMER with ETP~\cite{liu2024etp}, MERL~\cite{liu2024zero}, and C-MELT~\cite{c-melt} under conventional zero-shot settings across six datasets (Table~\ref{tab2}). On average, our method achieves an AUC of 76.00, surpassing the best baseline (MERL) by 0.75. These results demonstrate the strong representation learning capacity of CG-DMER and highlight the importance of semantic richness and contextual information in prompt-based zero-shot classification.
% Zero-shot classification using text prompts is a common task for assessing representation learning quality, which enables models to generalize to unseen tasks without requiring annotated data from those categories. In zero-shot learning, models typically rely on category names alone to make predictions, providing insufficient contextual information for accurate classification. However, by incorporating Large Language Models (LLMs), we can enhance the context by generating richer, clinically relevant descriptions of the categories. We use GPT-4o to customize the prompt for each category. We compare our method with ETP~\cite{liu2024etp}, MERL~\cite{liu2024zero}, and C-MELT~\cite{c-melt} under conventional zero-shot settings across six datasets, as summarized in Table~\ref{tab2}. On average, our method achieves an AUC of 76.00, outperforming the best baseline (MERL) by a margin of 0.75. This showcases the strong representation learning ability of our method and underscores the value of semantic richness and contextual information in prompt-based zero-shot classification.

\section{Ablation Studies and Analysis}
%This section provides extensive ablation studies on the key components of CG-DMER and reports the deep analysis of our method. We report the average performance of zero-shot classification and linear probing with 1\% data across six ECG classification datasets. Due to the page limit, we show more ablation studies and analysis in the Appendix.

\noindent
\textbf{Effects of Key Components}.  
We conduct ablation studies on the four core components of CG-DMER to evaluate their individual contributions (Table~\ref{tab4}). Variant~1 employs only the contrastive objective as a multimodal baseline. Incorporating spatial–temporal masked ECG modeling (Variant~2) improves AUC by +1.96 and +1.51, highlighting its effectiveness in capturing lead correlations and temporal dynamics. Variant~3 further adds masked text reconstruction, enabling the model to recover clinical semantics. Variant~4 introduces feature disentanglement, which mitigates modality bias and noise, yielding additional improvements. The full CG-DMER achieves the highest performance, demonstrating that all components are complementary and jointly enhance representation learning.
% \textbf{Effects of Key Components}. 
% We perform ablation studies on the four core components of CG-DMER to assess their contributions. As shown in Table~\ref{tab4}, Variant 1 uses only the contrastive objective as a multimodal baseline. Adding spatial-temporal masked ECG modeling (Variant 2) improves AUC by +1.96 and +1.51, showing its benefit for capturing lead correlations and temporal dynamics. Variant 3 adds masked text reconstruction to recover clinical semantics, while Variant 4 further applies feature decoupling to reduce modality bias and noise, yielding additional gains. The full CG-DMER achieves the best performance, confirming that all components are complementary enhance representation learning.

\begin{table}[t]
\caption{Effects of model components.}
\vspace{-5pt}
\centering
\label{tab4}
\resizebox{0.48\textwidth}{!}{
\renewcommand\arraystretch{1}
\begin{tabular}{c|cccc|cc}
\toprule[1.2pt]
Methods  & $\mathcal{L}_{e\_rec}$         & $\mathcal{L}_{t\_rec}$      & $\mathcal{L}_{orth}$   & $\mathcal{L}_{SigLIP}$    &       Linear (1\%)    &  Zero-shot \\
\midrule[1.2pt]
Variant 1                     & -            & -            & -            & -                     & 64.43                 & 73.36                                   \\
Variant 2                & \checkmark & -            & -            & -                      & 67.39                 & 74.87                                \\
Variant 3                & \checkmark & \checkmark & -            & -                      & 67.82                 & 75.00                                   \\
Variant 4                & \checkmark & \checkmark & \checkmark & -                       & 69.31                 & 75.77                                   \\                          CG-DMER                   & \checkmark & \checkmark & \checkmark & \checkmark &  \textbf{70.15}                 & \textbf{76.00}                                   \\
\bottomrule[1.2pt]

\end{tabular}}
\vspace{-15pt}
\end{table}

\begin{figure}[t]
\centering
    \includegraphics[width=0.96\linewidth]{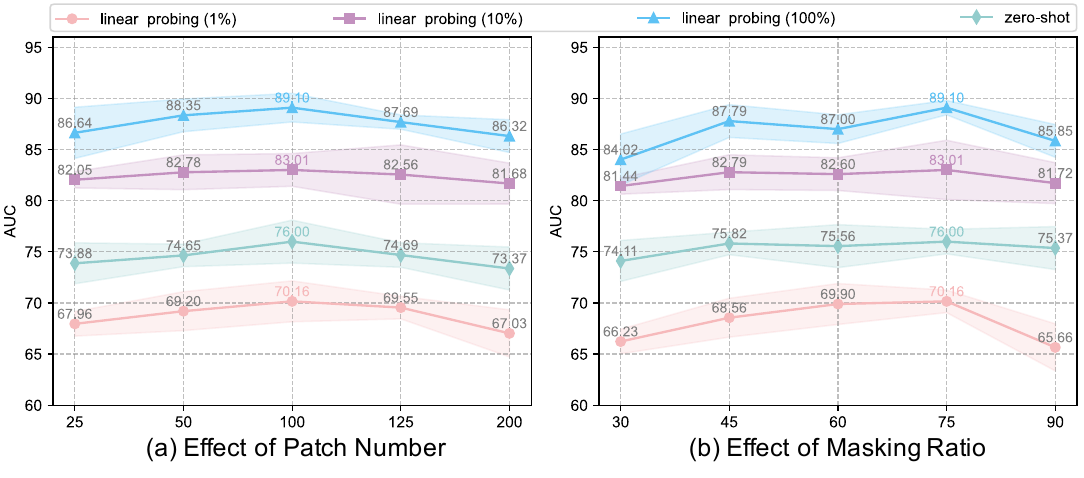} 

    \caption{Results of CG-DMER with different patch numbers and masking ratio on linear probing and zero-shot.}
    \vspace{-10pt}

    \label{fig3} 
\end{figure}

\noindent
\textbf{Patch Number and Masking Ratio}.  
Figure~\ref{fig3}(a) shows the effect of varying the patch number $N$, with the optimal value identified as $N=100$. Larger values (e.g., 200) degrade performance more severely than smaller ones (e.g., 25), likely because aggregating multiple segments into a single token introduces ambiguity.  
The masking ratio determines the proportion of unmasked patches provided to the decoder and thus controls the difficulty of the reconstruction task. As shown in Figure~\ref{fig3}(b), CG-DMER achieves its best performance at a moderately high masking ratio of 75\%.
% \textbf{Patch Number and Masking Ratio.} In Figure~\ref{fig3} (a), we ablate the patch number N and identify the optimal value as 100. A larger patch number (e.g., 200) has a more negative impact than smaller sizes (e.g., 25), likely because it converts multiple segments into one token, which introduces ambiguity. The masking ratio directly influences the number of unmasked patches available to the decoder, a factor that governs the complexity of the reconstruction task. Figure~\ref{fig3} (b) illustrates how performance changes with varying masking ratios. Notably, for CG-DMER, the highest performance is achieved at a moderately high masking ratio of 75\%.
%ecg序列长度是5000，划分成多少个patch 以及每条lead mask多少比例都会影响预训练的性能

\begin{figure}[t]
\centering
    \includegraphics[width=1\linewidth]{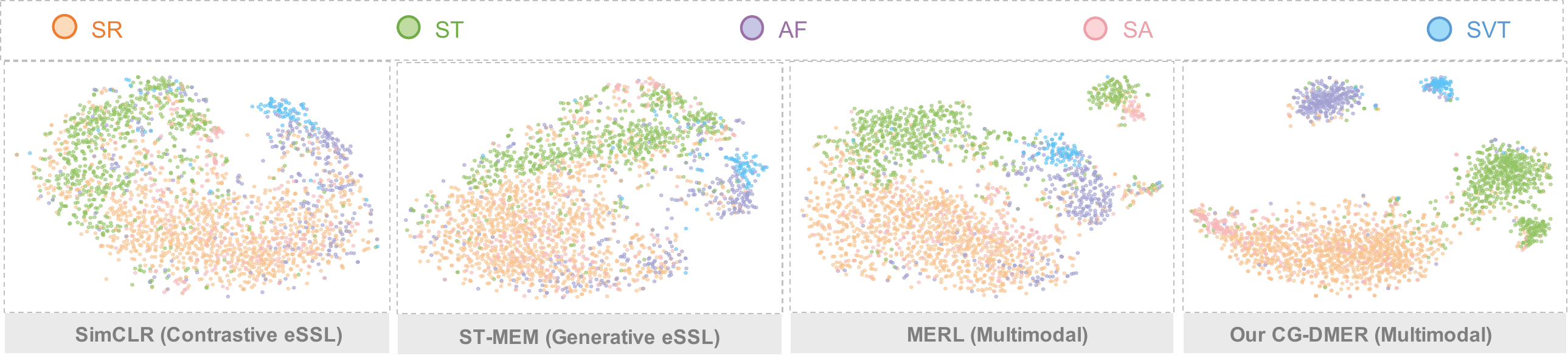} 

    \caption{T-SNE visualization on the CSN test set.}
    \vspace{-15pt}

    \label{fig5} 
\end{figure}

\noindent \textbf{Visualization of Learned ECG Representations}.  
To better understand the learned features, we visualize the final-layer outputs of the ECG encoder on the CSN test set using SimCLR, ST-MEM, MERL, and our CG-DMER. For clarity, samples from five diagnostic categories are shown. As illustrated in Figure~\ref{fig5}, multimodal methods separate categories even without supervision, whereas SimCLR and ST-MEM produce mixed and overlapping clusters. Compared to MERL, CG-DMER yields more compact and distinct clusters aligned with different ECG diagnoses, better reflecting clinical expectations.

\section{Conclusion}
In this paper, we present CG-DMER, a contrastive-generative framework for multimodal ECG representation learning. Using spatial-temporal masked modeling and representation disentanglement, CG-DMER captures fine-grained ECG dynamics and reduces modality-related issues. It enhances the discriminability and robustness of ECG representations, improving performance on various downstream tasks. Experiments on multiple datasets show that CG-DMER outperforms existing methods and has strong potential for clinical diagnostics.

\section{acknowledgments}
This work was supported by the Wellcome Leap’s Dynamic
Resilience Program jointly funded by Temasek Trust.

% \clearpage
\balance
\bibliographystyle{IEEEbib}
\bibliography{strings}

\begin{thebibliography}{10}

\bibitem{chen2020simple}
Ting Chen, Simon Kornblith, Mohammad Norouzi, and Geoffrey Hinton,
\newblock ``A simple framework for contrastive learning of visual representations,''
\newblock in {\em ICML}. PMLR, 2020, pp. 1597--1607.

\bibitem{mocov3}
X~Chen, S~Xie, and K~He,
\newblock ``An empirical study of training self-supervised vision transformers. in 2021 ieee,''
\newblock in {\em ICCV}, 2021, pp. 9620--9629.

\bibitem{astcl}
Ning Wang, Panpan Feng, Zhaoyang Ge, Yanjie Zhou, Bing Zhou, and Zongmin Wang,
\newblock ``Adversarial spatiotemporal contrastive learning for electrocardiogram signals,''
\newblock {\em IEEE Transactions on Neural Networks and Learning Systems}, 2023.

\bibitem{kiyasseh2021clocs}
Dani Kiyasseh, Tingting Zhu, and David~A Clifton,
\newblock ``Clocs: Contrastive learning of cardiac signals across space, time, and patients,''
\newblock in {\em ICML}. PMLR, 2021, pp. 5606--5615.

\bibitem{mckeen2024ecg}
Kaden McKeen, Laura Oliva, Sameer Masood, Augustin Toma, Barry Rubin, and Bo~Wang,
\newblock ``Ecg-fm: An open electrocardiogram foundation model,''
\newblock {\em arXiv preprint arXiv:2408.05178}, 2024.

\bibitem{hu2023spatiotemporal}
Rui Hu, Jie Chen, and Li~Zhou,
\newblock ``Spatiotemporal self-supervised representation learning from multi-lead ecg signals,''
\newblock {\em Biomedical Signal Processing and Control}, vol. 84, pp. 104772, 2023.

\bibitem{zhang2022maefe}
Huaicheng Zhang, Wenhan Liu, Jiguang Shi, Sheng Chang, Hao Wang, Jin He, and Qijun Huang,
\newblock ``Maefe: Masked autoencoders family of electrocardiogram for self-supervised pretraining and transfer learning,''
\newblock {\em IEEE Transactions on Instrumentation and Measurement}, vol. 72, pp. 1--15, 2022.

\bibitem{crt}
Wenrui Zhang, Ling Yang, Shijia Geng, and Shenda Hong,
\newblock ``Self-supervised time series representation learning via cross reconstruction transformer,''
\newblock {\em IEEE Transactions on Neural Networks and Learning Systems}, 2023.

\bibitem{na2024guiding}
Yeongyeon Na, Minje Park, Yunwon Tae, and Sunghoon Joo,
\newblock ``Guiding masked representation learning to capture spatio-temporal relationship of electrocardiogram,''
\newblock {\em arXiv preprint arXiv:2402.09450}, 2024.

\bibitem{liu2024etp}
Che Liu, Zhongwei Wan, Sibo Cheng, Mi~Zhang, and Rossella Arcucci,
\newblock ``Etp: Learning transferable ecg representations via ecg-text pre-training,''
\newblock in {\em ICASSP}. IEEE, 2024, pp. 8230--8234.

\bibitem{liu2024zero}
Che Liu, Zhongwei Wan, Cheng Ouyang, Anand Shah, Wenjia Bai, and Rossella Arcucci,
\newblock ``Zero-shot ecg classification with multimodal learning and test-time clinical knowledge enhancement,''
\newblock {\em arXiv preprint arXiv:2403.06659}, 2024.

\bibitem{c-melt}
Manh Pham, Aaqib Saeed, and Dong Ma,
\newblock ``C-melt: Contrastive enhanced masked auto-encoders for ecg-language pre-training,''
\newblock {\em arXiv preprint arXiv:2410.02131}, 2024.

\bibitem{MAE}
Kaiming He, Xinlei Chen, Saining Xie, Yanghao Li, Piotr Doll{\'a}r, and Ross Girshick,
\newblock ``Masked autoencoders are scalable vision learners,''
\newblock in {\em CVPR}, 2022, pp. 16000--16009.

\bibitem{hendrycks2016gaussian}
Dan Hendrycks and Kevin Gimpel,
\newblock ``Gaussian error linear units (gelus),''
\newblock {\em arXiv preprint arXiv:1606.08415}, 2016.

\bibitem{jin2023medcpt}
Qiao Jin, Won Kim, Qingyu Chen, Donald~C Comeau, Lana Yeganova, W~John Wilbur, and Zhiyong Lu,
\newblock ``Medcpt: Contrastive pre-trained transformers with large-scale pubmed search logs for zero-shot biomedical information retrieval,''
\newblock {\em Bioinformatics}, vol. 39, no. 11, pp. btad651, 2023.

\bibitem{zhai2023sigmoid}
Xiaohua Zhai, Basil Mustafa, Alexander Kolesnikov, and Lucas Beyer,
\newblock ``Sigmoid loss for language image pre-training,''
\newblock in {\em ICCV}, 2023, pp. 11975--11986.

\bibitem{gow2023mimic}
Brian Gow, Tom Pollard, Larry~A Nathanson, Alistair Johnson, Benjamin Moody, Chrystinne Fernandes, Nathaniel Greenbaum, Jonathan~W Waks, Parastou Eslami, Tanner Carbonati, et~al.,
\newblock ``Mimic-iv-ecg: Diagnostic electrocardiogram matched subset,''
\newblock {\em Type: dataset}, vol. 6, pp. 13--14, 2023.

\bibitem{simclr}
Ting Chen, Simon Kornblith, Mohammad Norouzi, and Geoffrey Hinton,
\newblock ``A simple framework for contrastive learning of visual representations,''
\newblock in {\em ICML}. PMLR, 2020, pp. 1597--1607.

\bibitem{byol}
Jean-Bastien Grill, Florian Strub, Florent Altch{\'e}, Corentin Tallec, Pierre Richemond, Elena Buchatskaya, Carl Doersch, Bernardo Avila~Pires, Zhaohan Guo, Mohammad Gheshlaghi~Azar, et~al.,
\newblock ``Bootstrap your own latent-a new approach to self-supervised learning,''
\newblock {\em NeurIPS}, vol. 33, pp. 21271--21284, 2020.

\bibitem{zbontar2021barlow}
Jure Zbontar, Li~Jing, Ishan Misra, Yann LeCun, and St{\'e}phane Deny,
\newblock ``Barlow twins: Self-supervised learning via redundancy reduction,''
\newblock in {\em ICML}. PMLR, 2021, pp. 12310--12320.

\bibitem{simsiam}
Xinlei Chen and Kaiming He,
\newblock ``Exploring simple siamese representation learning,''
\newblock in {\em CVPR}, 2021, pp. 15750--15758.

\bibitem{tstcc}
Emadeldeen Eldele, Mohamed Ragab, Zhenghua Chen, Min Wu, Chee~Keong Kwoh, Xiaoli Li, and Cuntai Guan,
\newblock ``Time-series representation learning via temporal and contextual contrasting,''
\newblock {\em arXiv preprint arXiv:2106.14112}, 2021.

\bibitem{clocs}
Dani Kiyasseh, Tingting Zhu, and David~A Clifton,
\newblock ``Clocs: Contrastive learning of cardiac signals across space, time, and patients,''
\newblock in {\em ICML}. PMLR, 2021, pp. 5606--5615.

\bibitem{wagner2020ptb}
Patrick Wagner, Nils Strodthoff, Ralf-Dieter Bousseljot, Dieter Kreiseler, Fatima~I Lunze, Wojciech Samek, and Tobias Schaeffter,
\newblock ``Ptb-xl, a large publicly available electrocardiography dataset,''
\newblock {\em Scientific data}, vol. 7, no. 1, pp. 1--15, 2020.

\bibitem{liu2018open}
Feifei Liu, Chengyu Liu, Lina Zhao, Xiangyu Zhang, Xiaoling Wu, Xiaoyan Xu, Yulin Liu, Caiyun Ma, Shoushui Wei, Zhiqiang He, et~al.,
\newblock ``An open access database for evaluating the algorithms of electrocardiogram rhythm and morphology abnormality detection,''
\newblock {\em Journal of Medical Imaging and Health Informatics}, vol. 8, no. 7, pp. 1368--1373, 2018.

\bibitem{zheng2020optimal}
Jianwei Zheng, Huimin Chu, Daniele Struppa, Jianming Zhang, Sir~Magdi Yacoub, Hesham El-Askary, Anthony Chang, Louis Ehwerhemuepha, Islam Abudayyeh, Alexander Barrett, et~al.,
\newblock ``Optimal multi-stage arrhythmia classification approach,''
\newblock {\em Scientific reports}, vol. 10, no. 1, pp. 2898, 2020.

\bibitem{zheng2022large}
Jianwei Zheng, Hangyuan Guo, and Huimin Chu,
\newblock ``A large scale 12-lead electrocardiogram database for arrhythmia study (version 1.0. 0),''
\newblock {\em PhysioNet}, 2022.

\end{thebibliography}

\end{document}